\documentclass[conference]{IEEEtran}
\usepackage{pifont}
\usepackage{times}

\usepackage[numbers]{natbib}
\usepackage{multicol}
\usepackage[bookmarks=true]{hyperref}

\usepackage{amsmath,amssymb}

\usepackage[table]{xcolor}

\usepackage{graphicx}
\usepackage{booktabs} %

\usepackage{mathtools}
\usepackage{algorithm}
\usepackage{tikz}
\usetikzlibrary{arrows,automata,positioning}
\usepackage{multirow} 
\usepackage{xspace}

\usepackage{comment}
\usepackage{amssymb} %
\usepackage{bbold}

\usepackage{tabulary}
\usepackage{tabularx}
\usepackage{makecell}
\usepackage{tabulary}
\usepackage{makecell}
\usepackage{wrapfig}
\usepackage{subcaption}

\usepackage{amssymb}
\usepackage{mathtools}
\usepackage{amsthm}
\usepackage{array}

\usepackage[capitalize,noabbrev]{cleveref}

\theoremstyle{plain}

\theoremstyle{definition}

\theoremstyle{remark}

\definecolor{nhred}{RGB}{178, 34, 34}
\newcommand{\algname}{Advanced Expressive Whole-Body Control\xspace}
\newcommand{\algabbr}{ExBody\textbf{\color{nhred}2}\xspace}

\definecolor{orange}{rgb}{1,0.5,0}
\definecolor{lightsalmonpink}{rgb}{1.0, 0.6, 0.6}
\definecolor{verylightsalmonpink}{rgb}{0.966, 0.805, 0.797}
\definecolor{lightblue}{rgb}{0.862, 0.906, 0.984}
\definecolor{lightyellow}{rgb}{1.0, 0.945, 0.797}
\definecolor{lightgreen}{rgb}{0.835, 0.91, 0.828}
\definecolor{lightpurple}{rgb}{0.879, 0.832, 0.902}

\newcommand{\fsize}{small}
\usepackage[font=\fsize,labelfont=bf,skip=5pt]{caption}
\usepackage[export]{adjustbox}
\usepackage{wrapfig}

\usepackage{ragged2e}

\usepackage{listings}

\definecolor{codegreen}{rgb}{0,0.6,0}
\definecolor{codegray}{rgb}{0.5,0.5,0.5}
\definecolor{codepurple}{rgb}{0.58,0,0.82}
\definecolor{backcolour}{rgb}{0.95,0.95,0.92}

\lstdefinestyle{mystyle}{
    backgroundcolor=\color{backcolour},   
    commentstyle=\color{codegreen},
    keywordstyle=\color{magenta},
    numberstyle=\tiny\color{codegray},
    stringstyle=\color{codepurple},
    basicstyle=\ttfamily\footnotesize,
    breakatwhitespace=false,         
    breaklines=true,                 
    captionpos=b,                    
    keepspaces=true,                 
    numbers=left,                    
    numbersep=5pt,                  
    showspaces=false,                
    showstringspaces=false,
    showtabs=false,                  
    tabsize=2
}

\usepackage{url}

\begin{document}

\title{\algabbr: Advanced Expressive Humanoid Whole-Body Control}

\author{
Mazeyu Ji$^{*,1}$ \quad \quad Xuanbin Peng$^{*,1}$ \quad \quad Fangchen Liu$^{2}$ \quad \quad Jialong Li$^{1}$
\vspace{0.1cm}\\
Ge Yang$^{3}$ \quad \quad Xuxin Cheng$^{\dagger,1}$ \quad \quad Xiaolong Wang$^{\dagger,1}$
\vspace{0.1cm}\\
$^1$UC San Diego \quad $^2$UC Berkeley \quad $^3$MIT\\
{\small $^*$Equal contribution $^\dagger$Equal advising} \\
\url{https://exbody2.github.io}
\vspace{-0.3cm} 
}

\twocolumn[{%
\renewcommand\twocolumn[1][]{#1}%
\maketitle

\begin{center}
    \centering
    \captionsetup{type=figure}
    \includegraphics[width=\linewidth]{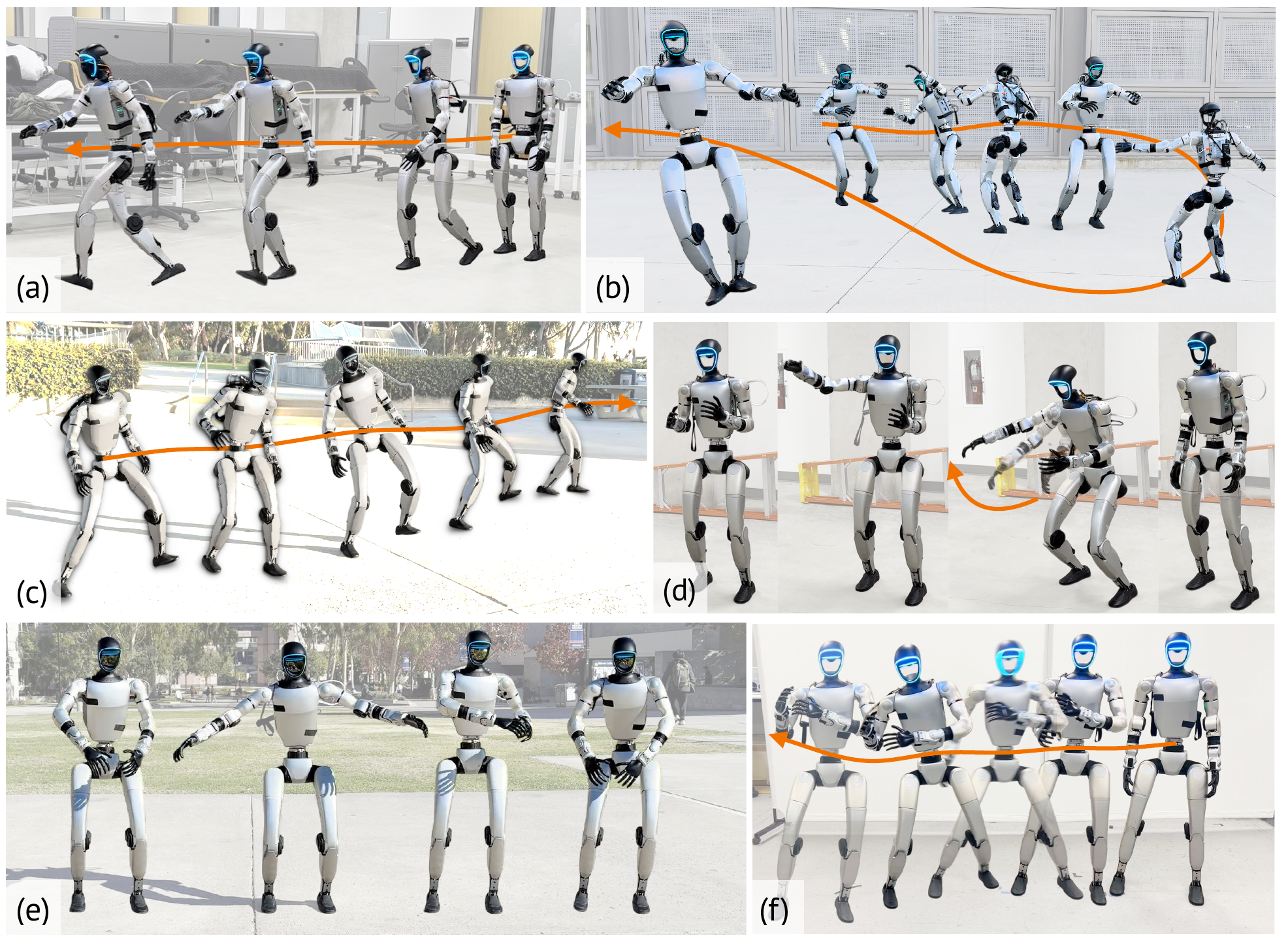}
    \caption{Humanoid robot executing various expressive whole-body motions in the real world. The robot can (a) walk with a large stride from static standing, (b) dance along a long horizon (43 seconds) choreography, (c) dynamic sidestep with fluid weight shifts, (d) punch with different height configurations, (e) express various upper-body movements while maintaining balance, (f) powerful rightward body hook with dynamic shifts.}
    \label{fig:teaser}
    \vspace{-0.0in}
\end{center}
}]

\IEEEpeerreviewmaketitle

\begin{abstract}
This paper tackles the challenge of enabling real-world humanoid robots to perform expressive and dynamic whole-body motions while maintaining overall stability and robustness. We propose \algname (Exbody2), a method for producing whole-body tracking controllers that are trained on both human motion capture and simulated data and then transferred to the real world. We introduce a technique for decoupling the velocity tracking of the entire body from tracking body landmarks. We use a teacher policy to produce intermediate data that better conforms to the robot's kinematics and to automatically filter away infeasible whole-body motions. This two-step approach enabled us to produce a student policy that can be deployed on the robot that can walk, crouch, and dance. We also provide insight into the trade-off between versatility and the tracking performance on specific motions. We observed significant improvement of tracking performance after fine-tuning on a small amount of data, at the expense of the others.

\end{abstract}

\section{Introduction}
\label{sec:intro}
The premise of humanoid robots is to enable human-like motions while occupying human living spaces. However, a humanoid robot with human-level \textit{expressiveness} and \textit{versatility} that is also robust in maintaining stability and control remains elusive. Inherent to this problem is the dynamic and kinematic gap between robots and biological body structures and the need for the controller to make a trade-off between expressiveness and stability. How to let robots imitate human whole-body motion across this gap and achieve both is a key challenge.

This paper introduces \algname (Exbody2), a framework that enables humanoid robots to perform expressive, human-like full-body motions with grace. At its core, Exbody2 features both a generalist and a specialist policy. The generalist policy, trained on diverse motion datasets, \textbf{outperforms previous approaches} by achieving high adaptability across a wide range of motions with a single policy. Building on this, we further fine-tune the policy for specific motion groups, producing specialist policies that ensure \textbf{even higher fidelity in targeted behaviors}. Our framework also incorporates a decoupled motion-velocity control strategy, allowing for both precise key body tracking and stable motion generation. These innovations set Exbody2 apart from previous methods and push the boundaries of humanoid motion control. Overall, there are three key innovations that distinguish Exbody2 from previous approaches:

(i) \textit{Generalist policy with automated data curation}. Human motion datasets often contain movements beyond a robot’s physical limits, making tracking difficult and reducing performance. Some methods refine datasets, like ExBody~\cite{cheng2024exbody} filtering motions via language labels, though ambiguous terms (e.g., “dance”) may still include infeasible actions. Others~\cite{he2024h20,he2024omnih2o} use SMPL avatars to simulate motions, but these can exceed real robot capabilities, impacting training. We identify the trade-off between dataset feasibility and diversity and develop an automated curation method that removes unsuitable lower-body motions while preserving diversity, enabling the policy to learn broad, expressive behaviors. Experiments validate that our method optimally balances feasibility and diversity, leading to improved stability and accuracy across diverse motion tasks.

(ii) \textit{Specialist policy with finetuning for targeted motions}. While the generalist policy enables broad motion coverage, finetuning enhances precision for specific motion groups. Motions with similar patterns are easier to learn under a shared policy, as they require consistent control strategies and constraints. Instead of training from scratch, we refine the generalist policy, leveraging its learned priors for efficient adaptation. This allows the policy to better capture fine-grained motion details and improve tracking accuracy for specialized tasks. Additionally, motion labels or an action recognition model can classify input motions, enabling dynamic selection of the most suitable specialist policy.

(iii) \textit{Decoupled motion-velocity control strategy}. Previous whole-body tracking approaches, such as H2O~\cite{he2024h20} and OmniH2O~\cite{he2024omnih2o}, rely on global tracking of keypoint positions. This often leads to tracking failures in immediate next steps when robots struggle to align with current global keypoints, limiting their applications to highly stationary scenarios. In contrast, Exbody2 converts keypoints into the local frame and decouples keypoint tracking from velocity control. Velocity-based tracking guides movement, while key body tracking focuses on motion imitation, emphasizing expressive motion reproduction. To improve tracking robustness, Exbody2 adopts a teacher-student framework, where a teacher policy is first trained using PPO~\cite{schulman2017proximal} with privileged information, including real root velocity, accurate body link positions, and physical properties (e.g., friction). The student policy is then trained via DAgger~\cite{dagger}-style distillation, using historical observations to infer privileged information, enabling real-world deployment without direct access to such data.

We evaluate our approach on the Unitree G1 against four state-of-the-art baselines, demonstrating higher fidelity across both simulation and real-world tests. In particular, the automated data curation yields a robust generalist policy that outperforms all previous methods as a single policy capable of handling diverse motions. Fine-tuning for specific tasks (e.g., dancing) further enhances motion quality and expressiveness. Overall, our results highlight ExBody2’s potential for bridging the gap between human-level expressiveness and reliable whole-body control in humanoid robots.

\section{Related Work}
\noindent\textbf{Humanoid Whole-Body Control.}
 Whole-body control for humanoid robots remains a complex and challenging problem due to the system’s high non-linearity and degrees of freedom. Traditional approaches predominantly rely on dynamics modeling and control \cite{miura1984dynamic, yin2007simbicon, hutter2016anymal, moro2019whole, dariush2008whole, kajita20013d, westervelt2003hybrid, kato1973development, hirai1998development, chignoli2021humanoid, dallard2023Sync, darvish2019wholebody, penco2019multimode, joao2019dynamic}, which often require accurate system identification and physical modeling, and intensive online computation for real-time control to handle different external perturbations for locomotion stability. Recent advances in reinforcement learning (RL) and sim-to-real transfer have demonstrated promising results in enabling complex whole-body skills for humanoid robots \cite{li2021reinforcement, li2023robust, siekmann2021blind, duan2023learning, li2024reinforcement, liao2024berkeley, TokenHumanoid2024, ito2022efficient, jeon2023learning, RealHumanoid2023, tang2023humanmimic, seo2023deep}. These approaches typically rely on training RL policies in simulation using task-specific rewards and environment randomization before transferring them to the real world. Notably, recent works such as \cite{cheng2024exbody, he2024h20, fu2024humanplus} have advanced real-world humanoid whole-body control for expressive motion by incorporating human motion datasets \cite{AMASS:2019} to guide RL training, with real-world applications such as motion imitation. However, these approaches still exhibit limitations in expressiveness and maneuverability, highlighting the untapped potential of humanoid robots. In contrast, our method enables the learning of more expressive and dynamic motions, enhancing the robot’s ability to perform complex whole-body movements.

\noindent\textbf{Robot Motion Imitation.}
Robot motion imitation can be broadly categorized into manipulation and locomotion areas. For manipulation tasks, robots are often wheeled or tabletop, prioritizing precise control over balancing and ground contact, making humanoid morphology unnecessary. Such robots can imitate the motion through direct teleoperation \cite{zhao2023learning, fu2024mobile, qin2022one}, portable devices \cite{wang2024dexcap, chen2024arcap, chi2024universal, ha2024umi} and learn from human videos with hand tracking or motion retargeting \cite{wang2023mimicplay, srirama2024hrp, li2024okami, kareer2024egomimicscalingimitationlearning}. In contrast, motion imitation for locomotion primarily aims to learn lifelike, natural behaviors from human or animal motion capture data. It requires precise control over contact dynamics, balance, and coordination across multiple degrees of freedom to achieve stable and realistic movement. While prior methods have enabled physics-based character motion imitation in simulation \cite{peng2021amp, 2022-TOG-ASE, tessler2023calm, InterPhysHassan2023, luo2024universal, ling2020character, zhang2023vid2player3d, wang2024strategy, tessler2024maskedmimic}, transferring diverse motions to real robots \cite{cheng2024exbody, he2024omnih2o, fu2024humanplus, he2024hover, RoboImitationPeng20, Escontrela22arXiv_AMP_in_real, dugar2024mhc} remains a significant challenge due to the hardware constraints. Previous methods \cite{cheng2024exbody, he2024omnih2o, fu2024humanplus, he2024hover} typically rely on manually filtering feasible motion data with human effort or hand-crafted heuristics. However, manually filtered datasets may still contain infeasible motions or lack diversity, limiting the robot's ability to fully utilize its hardware potential. Our method overcomes this challenge by automatically curating a diverse and feasible motion dataset, enabling more effective real-world deployment.

\begin{figure*}[htbp]
    \centering
    \includegraphics[width=\textwidth]{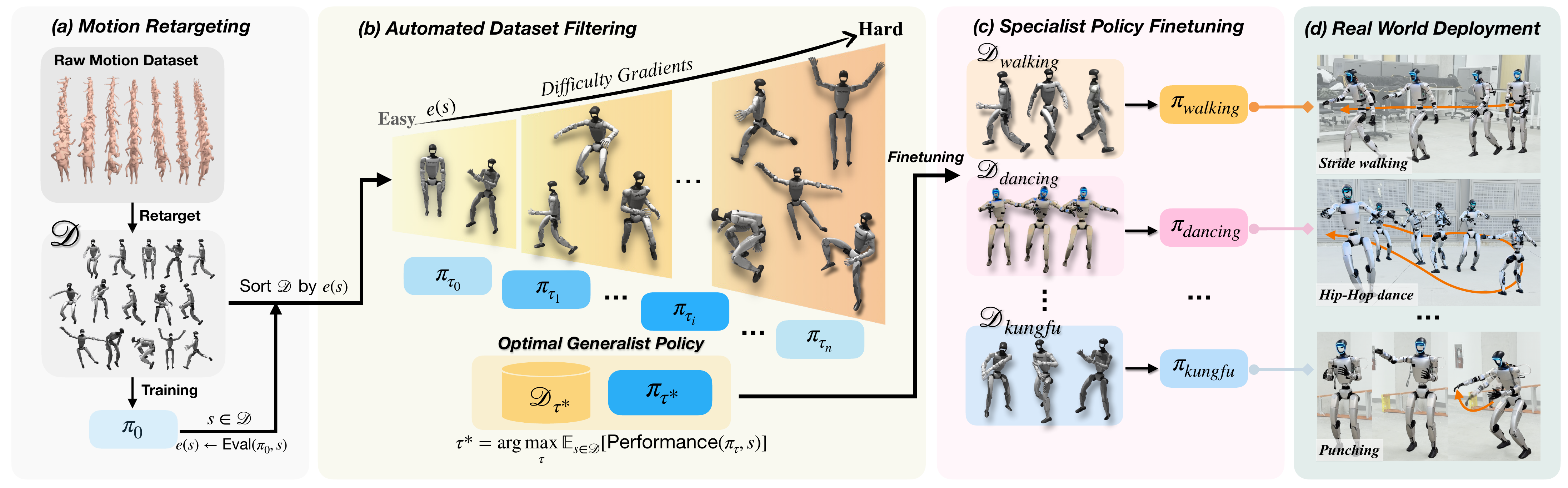}
    \caption{Exbody2's framework. (a) Motion retargeting adapts raw human motion datasets to fit the humanoid robot's morphology, generating a diverse set of training samples. (b) Automated dataset filtering ranks motions based on tracking errors and selects an optimal subset to train a generalist policy, balancing feasibility and diversity. (c) Specialist policy finetuning refines the generalist model for specific motion categories, such as walking, dancing, and kungfu, improving precision for targeted tasks. (d) The trained policies are deployed on a real humanoid robot, demonstrating expressive, dynamic, and stable whole-body motions in real-world environments.}

    \label{fig:main_figure}
\end{figure*}
\section{Exbody2: Learning Expressive Humanoid Whole-Body Control}
We propose \algname (Exbody2), a sim-to-real framework for expressive and robust whole-body control. As shown in Figure~\ref{fig:main_figure}, Exbody2 first retargets human motion data to fit the robot’s morphology, then trains a generalist policy using an automated dataset curation strategy to balance feasibility and diversity. To improve precision on specific motion groups, we further fine-tune specialist policies and deploy it onto real humanoid robots. In the following sections, we detail our generalist-specialist training pipeline, and our policy structure design, the two main contributions of our work.

\subsection{Data-driven Generalist-specialist Training Pipeline}

We adopt a Generalist--Specialist pipeline to balance adaptability and precision in whole-body motion tracking. 
At the heart of this pipeline is our \textbf{Feasibility--Diversity Principle}, which guides dataset design by requiring 
\emph{enough motion diversity} (particularly in the upper body) to cover a broad distribution of tasks, 
while maintaining \emph{feasibility} in the lower body to avoid unachievable or overly dynamic motions 
that degrade training stability. 
In practice, this amounts to filtering out extreme lower-body samples and retaining a wide range of upper-body actions. Further details on how we discover and validate this principle are in the appendix Section 3.

\noindent
\textbf{Generalist Policy.} 
Leveraging the Feasibility--Diversity Principle, we train a generalist policy on a broad dataset that has been 
automatically pruned to remove infeasible lower-body motions. 
This ensures that the policy learns expressiveness in upper-body movements 
without being hindered by physically impractical transitions.

\noindent
\textbf{Specialist Policies.} 
Once trained, the generalist policy can be fine-tuned for targeted tasks (e.g., dancing or kung fu), 
resulting in specialist policies with higher precision. 

This two-step approach balances overall adaptability with focused accuracy, 
as we demonstrate in both simulation and real-world evaluations.

\subsubsection{Generalist policy with automated data curation}
To obtain a policy \( \pi \) that performs well across diverse motion inputs, we first train an initial policy \( \pi_0 \) on a comprehensive, unfiltered motion dataset \( \mathcal{D} \), which is highly diverse with a lot of infeasible motions. After training \( \pi_0 \), we evaluate its tracking accuracy for each motion sequence \( s \in \mathcal{D} \), obtaining a tracking error metric \( e(s) \) that focuses on the lower body. The lower body plays a central role in dynamic feasibility and balance; thus, we focus on its tracking error for filtering. This preserves the upper body’s diversity while excluding excessively unstable motions, aligning with our feasibility--diversity principle. Specifically, we define
\[
e(s) \;=\; \alpha\, E_{\text{key}}(s) \;+\; \beta\, E_{\text{dof}}(s),
\]
where \( E_{\text{key}}(s) \) is the mean keybody position error for the lower body (preventing extreme deviations such as flipping or rolling), and \( E_{\text{dof}}(s) \) measures the mean joint-angle tracking error. The coefficients \( \alpha \) and \( \beta \) weight these two terms according to their relative importance for lower-body stability and precision. Once \( e(s) \) is computed for each sequence, we rank the motions by their tracking errors and derive the empirical distribution \( P(e) \).

The objective is to determine an error threshold \( \tau \) such that the subset of motion sequences with \( e(s) \leq \tau \), denoted as \( \mathcal{D}_\tau = \{ s \in \mathcal{D} \mid e(s) \leq \tau \} \), enables the training of a new policy \( \pi_\tau \) that maximizes performance across the full dataset \( \mathcal{D} \). Formally, we seek:  
\[
\tau^* = \arg\max_{\tau} \mathbb{E}_{s \in \mathcal{D}} [\text{Performance}(\pi_\tau, s)],
\]
where \( \pi_\tau \) is trained on \( \mathcal{D}_\tau \).  

In practice, we divide \( P(e)\) into evenly spaced error intervals to systematically evaluate the performance of policies trained on subsets corresponding to different thresholds \( \tau \). Although we use a greedy search to identify the optimal threshold \( \tau^* \), subsequent experiments reveal a strong trend in how the policy’s performance changes with \( \tau \). When \( \tau \) is too small, the filtered motions are overly simple, limiting the policy’s ability to generalize across the full dataset. Conversely, when \( \tau \) is too large, the inclusion of many infeasible motions introduces noise, degrading the training effectiveness. The best-performing policy is consistently obtained at a moderate \( \tau \), balancing diversity and feasibility.  

The optimal threshold \( \tau^* \), identified through this process, exhibits generalizability and can be effectively applied to other motion datasets, ensuring robust training and improved performance.  

\subsubsection{Specialist policy with finetuning for targeted motions}
After obtaining the generalist policy \( \pi_{\tau^*} \), which balances diversity and feasibility across a broad range of motions, we further refine it into a specialist policy tailored for specific, high-precision tasks. This finetuning process leverages the pretrained generalist policy rather than training a new policy from scratch, offering several advantages.  

First, specialist policies require tracking a smaller set of motions with higher precision, making finetuning a more efficient approach. Training directly from scratch for specialized motions is often impractical, as complex actions may be too difficult to learn without a strong prior. In contrast, the generalist policy provides a warm start, enabling the policy to adapt more effectively to challenging motions.  

Second, the generalist policy has been exposed to a wider range of motion sequences, improving robustness to variations and unexpected disturbances. While a policy trained from scratch on specific motions may perform well in simulation, it lacks exposure to diverse conditions. As a result, when deployed in the real world, such a policy may struggle with unseen disturbances or novel states, leading to failures. By finetuning from a generalist model, the specialist policy inherits adaptability and robustness, improving its real-world generalization.  

Finally, finetuning can reduce training time and computational requirements, as it refines an already well-trained model instead of learning from scratch. This makes it a practical and scalable approach for developing high-precision policies for specialized tasks while maintaining strong generalization capabilities.

\begin{figure}[t]
    \centering
    \includegraphics[width=\linewidth]{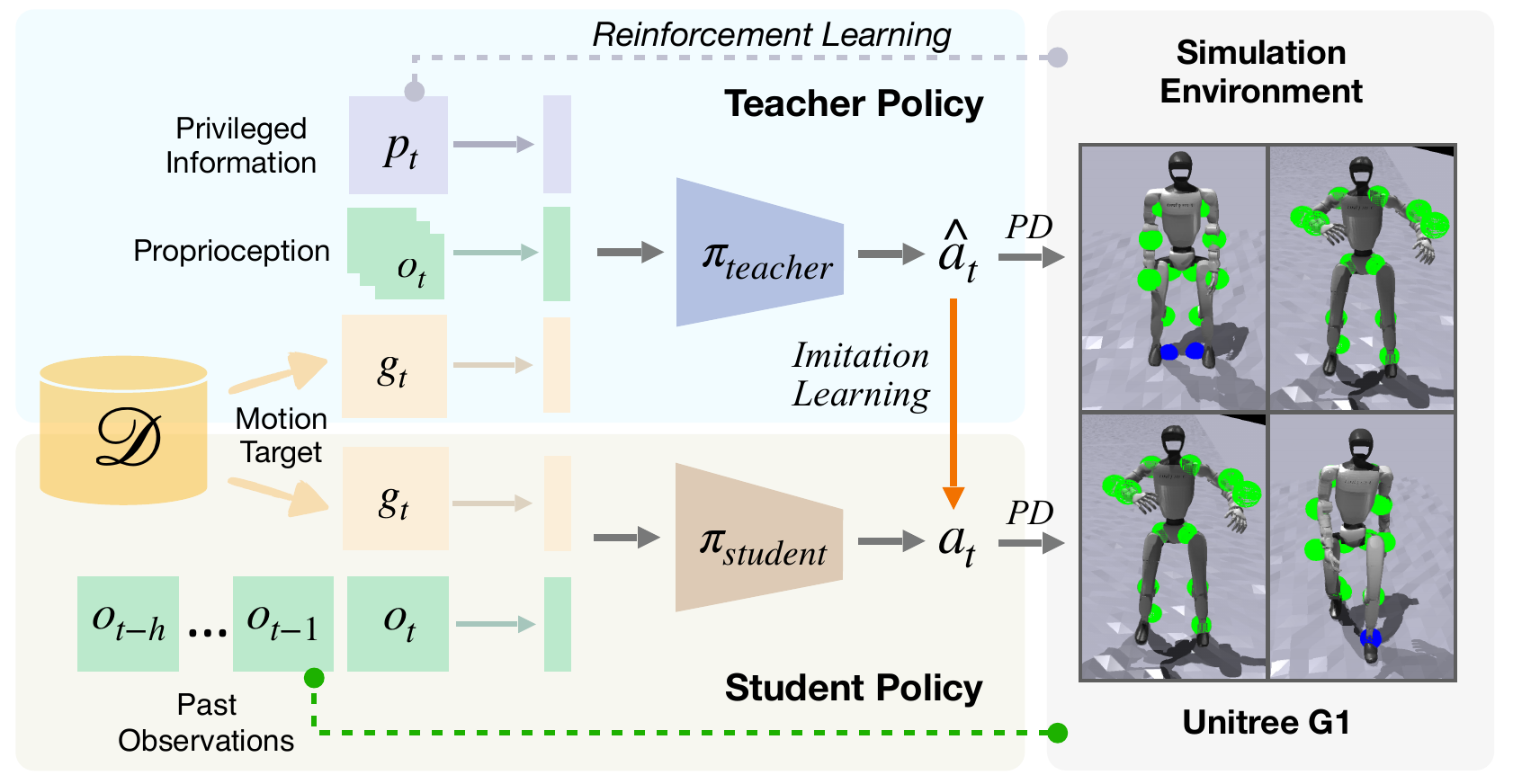}
    \caption{Teacher-student framework for humanoid motion learning, where the teacher uses privileged information, and the student learns from past observations to generate control actions.}
    \label{fig:architecture}
\end{figure}

\subsection{Policy Objective and Architecture}
Exbody2 aims at tracking a target motion more expressively in the whole body. To this end, Exbody2 adopts a two-stage teacher-student training procedure as in~\citep{lee2020learning, kumar2021rma}. Specifically, the oracle teacher policy is first trained with an off-the-shelf reinforcement learning (RL) algorithm, PPO~\cite{schulman2017proximal}, with privileged information that can be obtained only in simulators. For the second stage, we replace the privileged information with observations which are aligned with the real world, and distill the teacher policy to a deployable student policy. We train our policies using IsaacGym~\citep{makoviychuk2021isaac} with efficient parallel simulation. 
\subsubsection{Teacher Policy Training}
We can formulate the humanoid motion control problem as a \emph{Markov Decision Process} (MDP). The state space $\mathcal{S}$ contains privileged observation $\mathcal{X}$, proprioceptive states $\mathcal{O}$ and motion tracking target $\mathcal{G}$. A policy $\hat{\pi}$ takes $\{p_t, o_t, g_t\}$ as input, and outputs action $\hat{a}_t$, as illustrated in Figure~\ref{fig:architecture}, the teacher policy. The predicted action $\hat{a}_t \in R^{23}$ is the target joint positions of joint proportional derivative (PD) controllers. We use off-the-shelf PPO~\cite{schulman2017proximal} algorithm to maximize expectation of the accumulated future rewards $E_{\hat{\pi}}[\sum_{t=0}^{T}\gamma^{t}\mathcal{R}(s_t,\hat{a}_t)]$, which encourages tracking the demonstrations with robust behavior. The predicted $\hat{a}_t \in R^{23}$, which is the target position of joint proportional derivative (PD) controllers. \\
\textbf{Privileged Information.} The privileged information $p_t$ contains some ground-truth states of the humanoid robot and the environment, which can only be observed in simulators. It contains the ground-truth root velocity, real body links' positions, and physical properties (e.g. friction coefficients, motor strength). Privileged information can significantly improve the sample efficiency of RL algorithms, which is often leveraged to obtain a high-performing teacher policy. \\
\textbf{Motion Tracking Target.} Similar to Exbody~\citep{cheng2024express}, Exbody2 learns a policy that can be controlled by the joystick commands (e.g. the linear velocity and body pose) when accurately tracking a whole-body motion. The motion tracking target consists of two components, which are (1) the desired joints and 3D keypoints in both the upper and lower body and (2) target root velocity and root pose. 
For the full information about the privileged information, motion tracking information, and proprioceptive observations for the teacher policy, please refer to the supplementary materials. \\
\textbf{Reward Design.} Our reward function is carefully constructed to enhance the performance and realism of the humanoid robot's motion. The primary components of the reward include tracking the velocity, direction, and orientation of the root, alongside precise tracking of keypoints and joint positions. Additionally, we incorporate several regularization terms designed to boost the robot's stability and enhance the transferability from simulation to real-world applications. The main elements of our tracking reward are detailed in Table~\ref{tab:rewards_detailed}, while supplementary rewards aimed at improving stability and sim2real capabilities would discussed in the supplementary materials. \\
\subsubsection{Student Policy Training} 
In this stage, we remove the privileged information, and use longer history observation to train a student policy. As shown in Figure~\ref{fig:architecture}, the student policy encodes a series of past observations $o_{t-H:t}$ together with the encoded $g_t$ to get the predicted $a_t \sim \pi(\cdot|o_{t-H:t}, g_t)$. We supervise $\pi$ using the teacher's action $\hat{a}_t \sim \hat{\pi}(\cdot| o_t, g_t)$ with an MSE loss. 
$$l = \|a_t-\hat{a}_t\|^2$$
To train the student, we adopt the strategy used in DAgger~\cite{dagger}, we roll out the student policy $\pi$ in the simulation environment to generate training data. For each visited state, the teacher policy $\hat{\pi}$ computes the oracle action as the supervision signal. We proceed to refine the policy $\pi$ by iteratively minimizing the loss $l$ on the accumulated data. The training of $\hat{\pi}$ continues through successive rollouts until the loss $l$ reaches convergence. A critical aspect of training the student policy is preserving a sufficiently long sequence of historical observations. Detailed results and further analysis are elaborated in Section~\ref{sec:exp}.

\begin{table}[htbp!]
\centering
\normalsize
\begin{tabular}{@{}llr@{}}
\toprule
Term & Expression & Weight \\ \midrule
\multicolumn{3}{c}{Expression Goal $G^e$} \\ \midrule
DoF Position & exp$(-0.7 |\mathbf{q}_{\text{ref}}-\mathbf{q}|$) & 3.0 \\
                                  Keypoint Position & exp$(-|\mathbf{p}_{\text{ref}}-\mathbf{p}|)$ & 2.0 \\ \midrule
\multicolumn{3}{c}{Root Movement Goal $G^m$} \\ \midrule
                                  Linear Velocity & exp$(-4.0 |\mathbf{v}_\text{ref}-\mathbf{v}|)$ & 6.0 \\
                                  Velocity Direction & exp$(-4.0 \cos (\mathbf{v}_\text{ref}, \mathbf{v}))$ & 6.0 \\
                                  Roll \& Pitch & exp$(-|\boldsymbol{\Omega}_\text{ref}^{\phi\theta}-\boldsymbol{\Omega}^{\phi\theta}|)$ & 1.0 \\
                                  Yaw & exp$(-|\Delta y|)$ & 1.0 \\
\midrule
\end{tabular}
\caption{Rewards Specification for Exbody2.}
\label{tab:rewards_detailed}
\end{table}

\subsubsection{Motion-velocity Decoupled Control Strategy}
Motion tracking comprises two objectives: tracking DoF (joint) positions and keypoint (body keypoint) positions. Keypoint tracking usually plays a crucial role in tracking motions during training stage, as joint DoF errors can propagate to the whole body, while keypoint tracking is directly applied to the body. Existing work like H2O, OmniH2O~\cite{he2024h20, he2024omnih2o} learns to follow the trajectory of global keypoints. However, this global tracking strategy usually results in suboptimal or failed tracking behavior, as global keypoints may drift through time, resulting in cumulative errors that eventually hinder learning. To address this, we map global keypoints to the robot’s current coordinate frame, and instead utilize velocity-based global tracking. The coordination of velocity and motion allows tracking completion with maximal expressiveness, even if slight positional deviations arise. Moreover, to further enhance the robot's capabilities in following challenging keypoint motions, we allow a small global drift of keypoints during training stage and periodically correct them to the robot's current coordinate frame. During deployment, we strictly employ local keypoint tracking with motion-velocity decoupled control.

\section{Experiments}
\label{sec:exp}

In this section, we present experiments to evaluate the effectiveness of Exbody2. We first introduce the experimental setup and the baselines, followed by a detailed analysis addressing the following key research questions:

\noindent\textbf{Q1.} (Section \ref{sec:main_results}) Does Exbody2 generalist policy achieve higher tracking accuracy in both simulation and real-world deployment compared to prior methods?

\noindent\textbf{Q2.} (Section \ref{sec:data_curation}) What selection criteria lead to the optimal subset of a human motion dataset for learning a better generalist policy?

\noindent\textbf{Q3.} (Section \ref{sec:specialist}) Does finetuning a specialist policy for specific motion groups further improve tracking performance?

\begin{table*}[t]
\centering
\small
\begin{tabularx}{\textwidth}{l *{7}{X}}
\toprule
Method & $E_{\text{vel}} \downarrow$ & $E_{\text{mpkpe}} \downarrow$ & $E_{\text{mpkpe}}^{\text{upper}} \downarrow$ & $E_{\text{mpkpe}}^{\text{lower}} \downarrow$ & $E_{\text{mpjpe}} \downarrow$ & $E_{\text{mpjpe}}^{\text{upper}} \downarrow$ & $E_{\text{mpjpe}}^{\text{lower}} \downarrow$ \\
\midrule
Exbody  & 0.4700 & 0.1339 & 0.1249 & 0.1428 & 0.2020 & 0.1343 & 0.2952 \\
Exbody\textsuperscript{\( \dagger \)}  & 0.4195 & 0.1150 & 0.1106 & 0.1198 & 0.1496 & 0.1416 & 0.1607 \\
OmniH2O*  & 0.3725 & 0.1253 & 0.1266 & 0.1240 & 0.1681 & 0.1564 & 0.1843 \\
Exbody2-w/o-Filter & \textbf{0.2787} & 0.1133 & 0.1087 & 0.1182 & 0.1355 & 0.1192 & 0.1579 \\
\textbf{Exbody2(Ours)}  & 0.2930 & \textbf{0.1000} & \textbf{0.0960} & \textbf{0.1040} & \textbf{0.1079} & \textbf{0.0953} & \textbf{0.1253} \\
\midrule
\end{tabularx}
\caption{Comparisons with baselines on dataset $\mathcal{D}_{CMU}$ for Unitree G1. We loop each motion trajectory 10 times in simulation and compute the per-step average error for each policy over all repetitions. The lowest error in each column for each group is highlighted in bold.}
\label{tab:baseline}
\end{table*}

\begin{table}[t]
\centering
\small
\begin{tabular}{l|ccc}
\toprule
\textbf{Method} & 
$E_{\text{mpjpe}} \downarrow$ & 
$E_{\text{mpjpe}}^{\text{upper}} \downarrow$ &
$E_{\text{mpjpe}}^{\text{lower}} \downarrow$ \\
\midrule
Exbody & 0.2178 & 0.1223 & 0.3239\\
Exbody\textsuperscript{\( \dagger \)} & 0.1465 & 0.1314 & 0.1672\\
OmniH2O* & 0.1396 & 0.1273 & 0.1533\\
Exbody2-w/o-Filter & 0.1361 & 0.1254 & 0.1481\\
Exbody2(Ours) & \textbf{0.1074} & \textbf{0.1092} & \textbf{0.1054}\\
\bottomrule
\end{tabular}
\caption{Comparisons with baselines on selected motions for Unitree G1 in real world.}
\label{tab:realrobot}
\end{table}

\subsection{Experimental Setup}

\noindent \textbf{Baselines.} To examine the effectiveness of varying tracking methods, motion control strategies, and training techniques, we evaluate four baselines using a classic CMU dataset~\cite{cmu_mocap}, which includes a wide variety of action types.
\begin{itemize}
    \item \textbf{Exbody}~\cite{cheng2024express}: This method utilizes one-stage RL training pipeline, and only tracks the upper body movements from the human data, while tracking the root motion of the lower body without explicitly following step patterns and focusing on partial body tracking.

    \item \textbf{Exbody\textsuperscript{\( \dagger \)}}: The whole-body control version of Exbody, where the full-body movements are tracked based on human data. This setup enables comprehensive human motion imitation, attempting to match the entire body posture to the reference data.

    \item \textbf{OmniH2O*}: Our reproduction of the OmniH2O~\cite{he2024omnih2o}, using global keypoints tracking and the same observation space as described in the original paper. During deployment, we adapt OmniH2O to our local tracking evaluation for fair comparison.

    \item \textbf{Exbody2}: Our method utilizes local keypoint tracking and employs data curation to refine the training set. It incorporates various training techniques to enhance overall motion fidelity and improve sim-to-real transfer.

\end{itemize}

\noindent \textbf{Metrics.}
We evaluate the policy's performance using several metrics calculated across all motion sequences from the dataset. The \emph{mean linear velocity error} $E_{\text{vel}}$ (m/s) measures the error between the robot's root linear velocity and that of the demonstration, indicating the policy's ability to track velocity. We calculate tracking error in terms of keybody positions and joint angles. The \emph{Mean Per Keybody Position Error} (MPKPE) $E_{\text{mpkpe}}$ (m) evaluates the overall keypoint position tracking ability. For more detailed analysis, we report the MPKPE for the upper body $E_{\text{mpkpe}}^{\text{upper}}$ (m) and the lower body $E_{\text{mpkpe}}^{\text{lower}}$ (m), assessing the keypoint position tracking of the upper and lower body, respectively. Similarly, the \emph{Mean Per Joint Position Error} (MPJPE) $E_{\text{mpjpe}}$ (rad) measures the joint tracking ability. We also report the MPJPE for the upper body $E_{\text{mpjpe}}^{\text{upper}}$ (rad) and the lower body $E_{\text{mpjpe}}^{\text{lower}}$ (rad) to evaluate tracking performance in different body regions.

More details about environment designs and baseline implementations can be found in the supplementary material.

\subsection{Generalist Policy Performance}
\label{sec:main_results}

As shown in Table \ref{tab:baseline}, our \emph{Exbody2} policy already surpasses prior baselines (\emph{Exbody}, \emph{Exbody\textsuperscript{\( \dagger \)}}, and \emph{OmniH2O*}) across all reported metrics in simulation without motion filter when trained on the complete dataset. By further incorporating motion filtering, \emph{Exbody2} achieves additional gains in both upper- and lower-body tracking. The improvement in lower-body accuracy is especially notable, leading to greater global stability and, consequently, more precise upper-body control.

The only trade-off observed is a slight increase in velocity tracking error compared to the unfiltered version. We attribute this to the broader velocity patterns found in the full dataset, which facilitate more diverse dynamic behaviors but also introduce additional noise. Overall, the modest velocity trade-off is outweighed by significant enhancements in stability and precision. These trends are observed under simulation conditions, and we further analyze real-world performance in the following section.

In real-world experiments, we selected a representative subset from the CMU dataset, encompassing a diverse range of actions—such as expressive standing postures, walking at various speeds and patterns, squatting, and dancing. Using the robot’s onboard motor readings, we computed the joint position error, as reported in Table~\ref{tab:realrobot}. The experimental outcomes closely mirror the simulation results, demonstrating that our algorithm achieves higher tracking accuracy compared to other baselines for both upper and lower body segments. Notably, the integration of automated data curation has substantially enhanced performance. This improvement is critical in real-world environments, where unanticipated disturbances are more prevalent, underscoring the importance of maintaining robust and consistent behavior to achieve high-precision tracking throughout the motion sequences.

Overall, our generalist policy achieves significant improvements in full-body tracking accuracy for both upper and lower body, and velocity tracking accuracy compared to the baseline algorithms both in simulation and the real world, demonstrating stable and effective tracking performance in dynamic environments.

\subsection{Impact of Automatic Data Curation}
\label{sec:data_curation}

\begin{figure}[t]
    \centering
    \includegraphics[width=\linewidth]{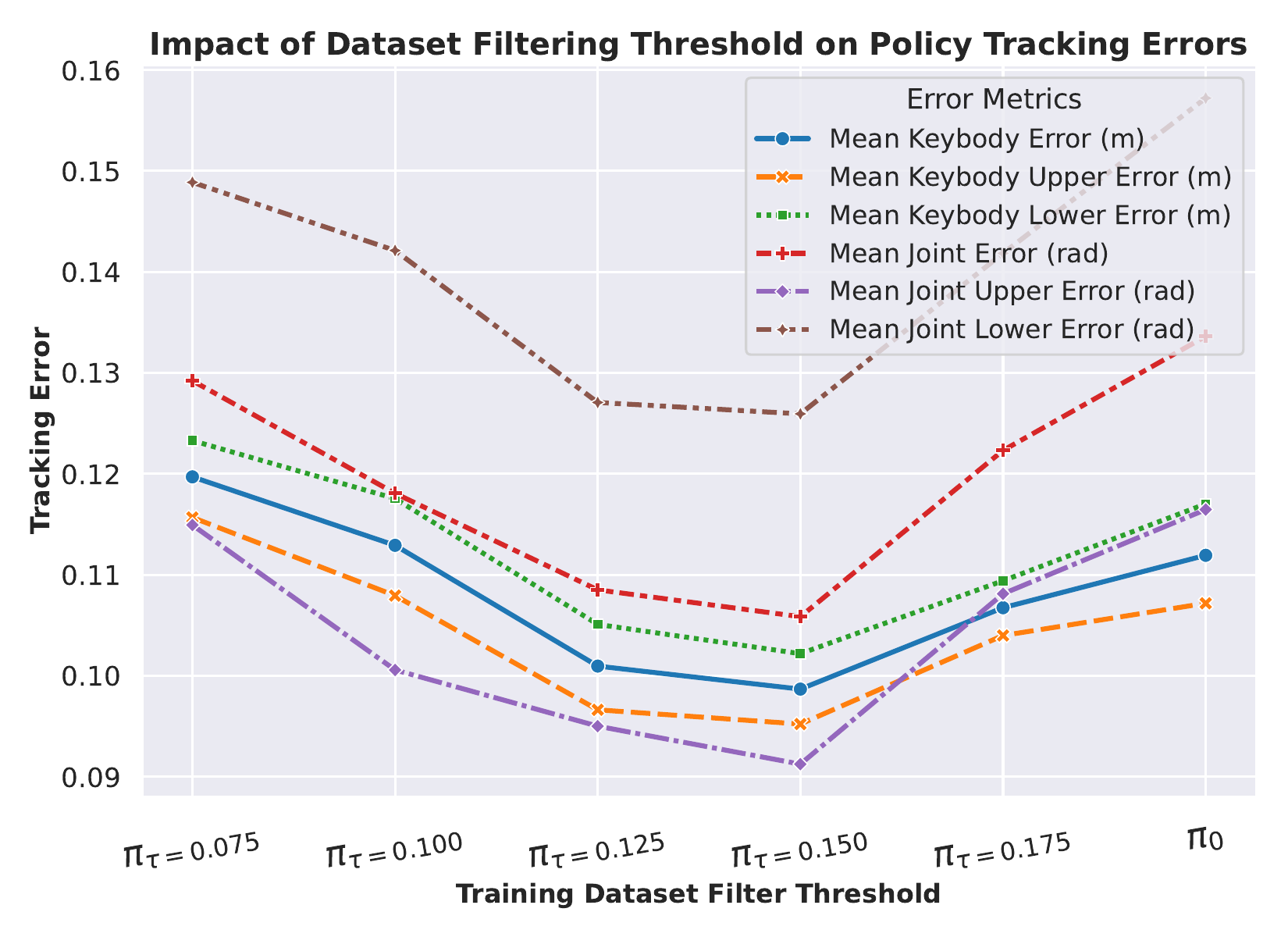}
    \caption{Impact of dataset filtering thresholds on policy tracking errors. 
    The figure shows the tracking error trends across different dataset filtering thresholds. Policies trained on datasets with filtering thresholds that balance diversity and stability (e.g., \( \pi_{\tau=0.150} \)) achieve the lowest tracking errors. The base policy exhibits suboptimal performance due to unfiltered data, while overly restrictive thresholds (e.g., \( \pi_{\tau=0.075} \)) and overly lenient thresholds (e.g., \( \pi_{\tau=0.175} \)) show reduced effectiveness. We compute the error metric \(e(s) = \alpha\, E_{\text{key}}(s) + \beta\, E_{\text{dof}}(s)\) with \(\alpha=0.1, \beta=0.9\), 
    assigning heavier weight to the joint-angle term.}
    \label{fig:tracking_errors}
\end{figure}

To experimentally examine the selection criteria for constructing an optimal human motion dataset that enhances generalist policy learning, we conducted the following experiment, reconstructing the full pipeline of our automated data curation method and evaluating its effect on policy performance.

\begin{enumerate}
    \item \textbf{Base Policy Training}: We first trained a base policy on the unfiltered $\mathcal{D}_{CMU}$ dataset, which encompasses a broad range of human motions, including both stable and highly dynamic sequences, as well as infeasible movements that exceed the robot’s physical limitations.

    \item \textbf{Dataset Filtering}: The base policy's tracking performance was evaluated on $\mathcal{D}_{CMU}$, and each motion sequence was assigned a tracking error score. Based on these scores, we sorted the sequences and applied filtering thresholds to create progressively refined datasets: $ \mathcal{D}_{\tau=0.075}$, $\mathcal{D}_{\tau=0.1}$, $\mathcal{D}_{\tau=0.125}$, $\mathcal{D}_{\tau=0.15}$, $\mathcal{D}_{\tau=0.175}$. A lower threshold retains only motions with the lowest tracking errors, ensuring high stability but reducing diversity. Conversely, a higher threshold allows more challenging and diverse motions but introduces tracking instability.
    
    These thresholds were determined based on the error distribution of the base policy, as detailed in the appendix Section 4.B. The distribution-informed selection ensures that the filtering process is not arbitrary but rather grounded in empirical data. This selection strategy is not limited to a specific dataset; rather, it generalizes to other datasets with similar policy architectures and training conditions. When applied to new datasets, it effectively filters motions while maintaining a balance between feasibility and diversity.
    
    \item \textbf{Filtered Policy Resume:} 
For each filtered dataset $\mathcal{D}_{\tau}$, we resume training from the base policy $\pi_0$ to obtain a new policy $\pi_{\tau}$. 
Concretely, thresholds $\tau \in \{0.075, 0.1, 0.125, 0.15, 0.175\}$ each yield a distinct subset $\mathcal{D}_{\tau}$, 
and thus a correspondingly refined policy $\pi_{\tau}$. 
This resumption process leverages the base model’s learned prior, enabling faster adaptation to each subset’s motion characteristics 
and improving overall training efficiency.

    \item \textbf{Evaluation}: All the resulting policies were evaluated on the full $\mathcal{D}_{CMU}$ dataset, measuring tracking performance across multiple metrics. The results are visualized in Figure \ref{fig:tracking_errors}.
    
\end{enumerate}

The results confirm that dataset selection significantly influences both generalization and tracking accuracy. 
A well-chosen dataset, striking a balance between stability and diversity, produces more robust and adaptable policies, 
in line with our \emph{feasibility-diversity principle}.

\begin{itemize}
    \item \textbf{Low Thresholds} (e.g., $ \pi_{\tau=0.075} $): Policies trained on heavily filtered datasets exhibited poor generalization, as they primarily learned from static and simple motions. While these datasets ensured stability, the lack of diversity limited the policy’s ability to handle more complex behaviors.
    
    \item \textbf{High Thresholds} (e.g., $ \pi_{\tau=0.175}, \pi_{0} $): Policies trained on datasets with high thresholds struggled due to the inclusion of highly dynamic and unstable motions. The increased data variability introduced noise during training, leading to inconsistent policy behavior and reduced tracking accuracy.
    
    \item \textbf{Moderate Thresholds} (e.g., $ \pi_{\tau=0.15} $): Policies trained on datasets with intermediate thresholds achieved the best trade-off between feasibility and diversity. The dataset $ \mathcal{D}_{\tau=0.15} $ retained sufficient variability to improve generalization while excluding excessively difficult or unstable motions, leading to the lowest overall tracking error.
\end{itemize}

These findings validate the importance of carefully balancing feasibility and diversity in dataset selection. The dataset $ \mathcal{D}_{\tau=0.15} $ and its corresponding policy $ \pi_{\tau=0.15} $ emerged as the optimal choice, ensuring both strong generalization and stable tracking performance. This highlights the necessity of structured dataset curation in developing robust robotic policies, particularly for real-world deployment where adaptability and reliability are crucial.

\subsection{Specialist Policy finetuning}
\label{sec:specialist}

\begin{table*}[t]
\centering
\small
\begin{tabularx}{\textwidth}{l *{7}{X}}
\toprule
\textbf{Method} & $E_{\text{vel}}\downarrow$ & $E_{\text{mpkpe}}\downarrow$ & $E_{\text{mpkpe}}^{\text{upper}}\downarrow$ & $E_{\text{mpkpe}}^{\text{lower}}\downarrow$ & $E_{\text{mpjpe}}\downarrow$ & $E_{\text{mpjpe}}^{\text{upper}}\downarrow$ & $E_{\text{mpjpe}}^{\text{lower}}\downarrow$ \\
\midrule
\rowcolor{lightgray}
\multicolumn{8}{l}{\textbf{(b) $\mathcal{D}_{easy}$}} \\
Specialist
& \textbf{0.0828} 
& \textbf{0.0561} 
& \textbf{0.0564} 
& \textbf{0.0558} 
& \textbf{0.0772} 
& \textbf{0.0647} 
& \textbf{0.0944} \\
Scratch
& 0.0853 
& 0.0608 
& 0.0623 
& 0.0592 
& 0.0843 
& 0.0711 
& 0.1024 \\
Generalist
& 0.0986 
& 0.0699 
& 0.0708 
& 0.0690 
& 0.1041 
& 0.0882 
& 0.1259 \\
\midrule

\rowcolor{lightgray}
\multicolumn{8}{l}{\textbf{(a) $\mathcal{D}_{Moderate}$}} \\
Specialist
& \textbf{0.0991} 
& \textbf{0.0571} 
& \textbf{0.0582} 
& \textbf{0.0559} 
& \textbf{0.0760} 
& \textbf{0.0636} 
& \textbf{0.0930} \\
Scratch
& 0.1188 
& 0.0676 
& 0.0688 
& 0.0663 
& 0.0924 
& 0.0794 
& 0.1103 \\
Generalist
& 0.1217 
& 0.0741 
& 0.0727 
& 0.0755 
& 0.1092 
& 0.0914 
& 0.1337 \\
\midrule

\rowcolor{lightgray}
\multicolumn{8}{l}{\textbf{(c) $\mathcal{D}_{Hard}$}} \\
Specialist
& 0.1712 
& \textbf{0.0827} 
& \textbf{0.0829} 
& \textbf{0.0826} 
& \textbf{0.1047} 
& \textbf{0.0911} 
& \textbf{0.1234} \\
Scratch
& 0.1631 
& 0.0886 
& 0.0898 
& 0.0873 
& 0.1188 
& 0.1067 
& 0.1354 \\
Generalist
& \textbf{0.1452} 
& 0.0890 
& 0.0867 
& 0.0912 
& 0.1181 
& 0.1011 
& 0.1414 \\
\midrule

\rowcolor{lightgray}
\multicolumn{8}{l}{\textbf{(d) $\mathcal{D}_{ACCAD}$}} \\
Specialist
& 0.4021 
& \textbf{0.1149} 
& \textbf{0.1079} 
& \textbf{0.1215} 
& \textbf{0.1402} 
& \textbf{0.1290} 
& \textbf{0.1557} \\
Scratch
& 0.4153 
& 0.1246 
& 0.1154 
& 0.1332 
& 0.1609 
& 0.1490 
& 0.1771 \\
Generalist
& \textbf{0.3361} 
& 0.1268 
& 0.1156 
& 0.1391 
& 0.1716 
& 0.1532 
& 0.1967 \\
\bottomrule
\end{tabularx}
\caption{Comparison of three training strategies (\emph{Generalist}, \emph{Specialist}, \emph{Scratch}) across four dataset groups ($\mathcal{D}_{easy}$, $\mathcal{D}_{moderate}$, $\mathcal{D}_{hard}$, and $\mathcal{D}_{ACCAD}$).}
\label{tab:groups_comparison}
\end{table*}

\begin{figure}[t]
    \centering
    \includegraphics[width=\linewidth]{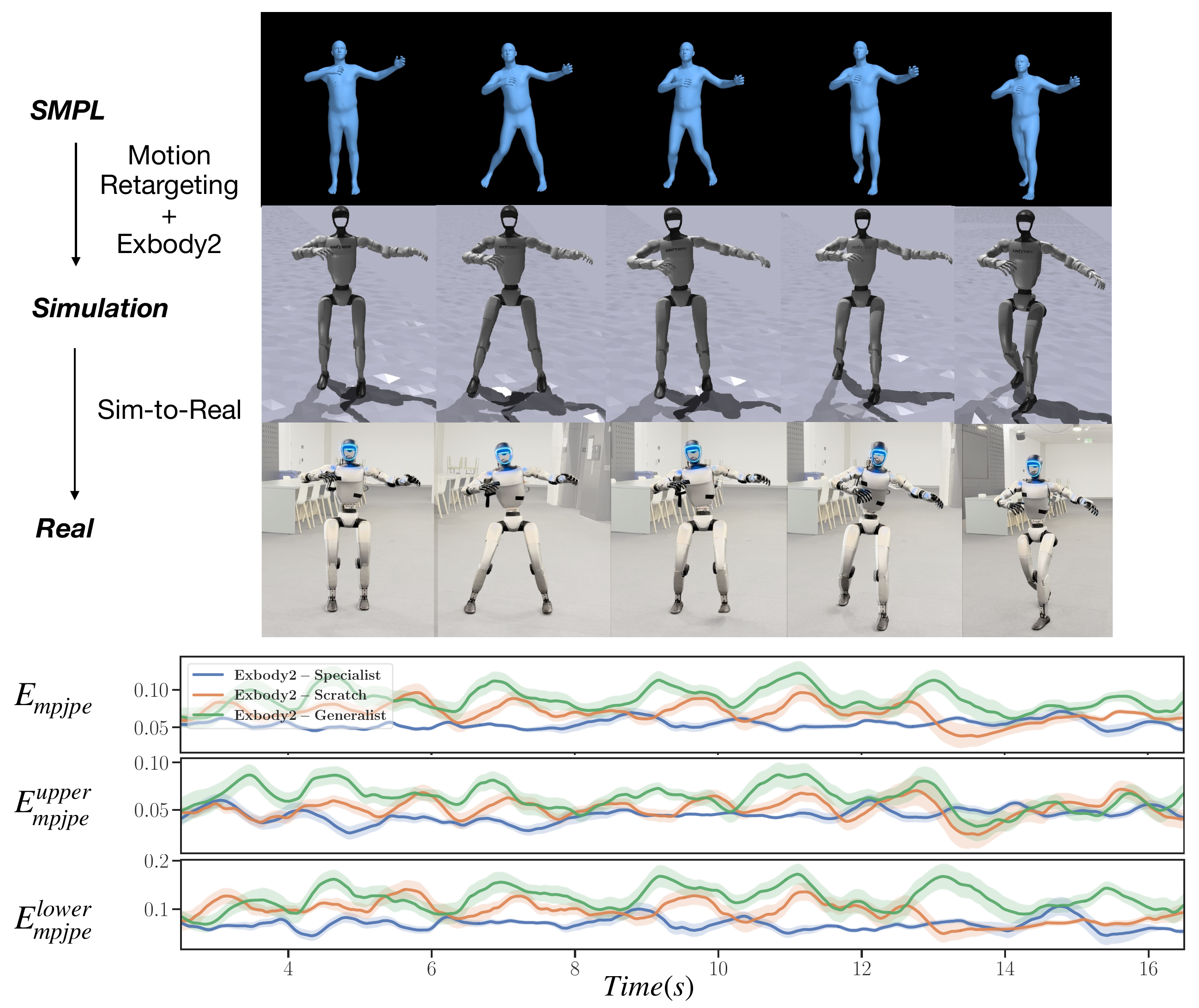}
    \caption{A sequence of a robot performing the Cha-Cha dance. From top to bottom: the reference motion represented by an avatar, our algorithm's performance in the simulation, and its performance on a real robot. The bottom three rows show the per-frame errors: whole-body joint DoF error, upper-body joint DoF error, and lower-body DoF error, with the blue curve representing \emph{Exbody2-Specialist} policy finetuned on $\mathcal{D}_{dancing}$ , orange for \emph{Exbody2-Scratch} policy training from scratch on $\mathcal{D}_{dancing}$, green for our \emph{Exbody2-Generalist} policy trained on filtered $\mathcal{D}_{CMU}$.}
    \label{fig:chacha}
\end{figure}

We evaluate the effectiveness of the pretrain-finetune paradigm in enhancing policy performance. To this end, we compare three distinct training strategies:

\begin{itemize}
    \item \textbf{Generalist}: A generalist policy trained on the automatically filtered dataset (\( \pi_{\tau=0.15} \)), designed to provide broad motion coverage and strong generalization across diverse tasks.
    \item \textbf{Specialist}: A specialist policy obtained by fine-tuning the pretrained generalist policy on task-specific datasets, enabling higher precision for specialized motions.
    \item \textbf{Scratch}: A policy trained from scratch on the same task-specific datasets. To ensure fairness, we match its total training iterations to the sum of \emph{pretraining} plus \emph{fine-tuning} iterations used in the Specialist approach.
\end{itemize}

To assess the robustness and adaptability of these policies, we conduct experiments across four manually curated datasets:

\begin{itemize}
    \item \( \mathcal{D}_{\text{easy}}, \mathcal{D}_{\text{moderate}}, \mathcal{D}_{\text{hard}} \): A series of datasets with increasing difficulty levels, categorized based on motion dynamics. Lower-difficulty datasets mainly contain static or low-movement motions, while higher-difficulty datasets include more dynamic and high-momentum movements. This progression allows us to assess how well the policies generalize to increasingly complex motions.
    
    \item \( \mathcal{D}_{\text{ACCAD}} \): An out-of-distribution (OOD) dataset used to evaluate the generalization capability of the learned policies on previously unseen motion patterns.
\end{itemize}

The quantitative evaluation results are summarized in Table~\ref{tab:groups_comparison}. The key findings from these experiments are discussed below.

\begin{enumerate}
    \item \textbf{Performance on \( \mathcal{D}_{\text{easy}}, \mathcal{D}_{\text{moderate}}, \mathcal{D}_{\text{hard}} \)}: 
    The finetuned policy achieves the best performance across all difficulty levels. The more challenging the dataset, the greater the advantage of finetuning over training from scratch, demonstrating the importance of leveraging a pretrained policy as a foundation for specialized tasks. For highly dynamic motions, the generalist policy exhibits slightly better velocity tracking due to its broader exposure to diverse movements. However, the specialist policy consistently achieves higher overall precision.
    
    \item \textbf{Performance on \( \mathcal{D}_{\text{ACCAD}} \)}: 
    The finetuned policy significantly outperforms both the pretrained and scratch policies on the OOD dataset, highlighting its superior generalizability and adaptability to unseen scenarios. This result further confirms that the pretrained generalist policy provides a strong foundation, while finetuning enhances task-specific adaptation.
\end{enumerate}

To better illustrate the effectiveness of the specialist policy, we select the Cha-Cha dance as a case study, as shown in Figure~\ref{fig:chacha}. Cha-Cha is one of the representative motions in the dance group and involves dynamic lower-body movements combined with expressive upper-body gestures. By comparing policies trained from scratch, the generalist policy, and the specialist policy fine-tuned on the dancing dataset, we observe that the specialist policy achieves significantly lower tracking errors across all key metrics. This result further confirms the advantage of task-specific finetuning in capturing fine-grained motion details while maintaining stability.

In conclusion, the pretrain-finetune paradigm proves to be an effective strategy for achieving robust and adaptable policies. The pretrained generalist policy (\( \pi_{\tau=0.15} \)) provides a strong starting point, while finetuning allows specialization for specific tasks, resulting in superior performance across diverse datasets. This approach demonstrates significant benefits, particularly in challenging and OOD scenarios, and highlights the importance of combining generalist capabilities with task-specific specialization.

\section{Conclusion}
This paper introduces \algname (Exbody2), a novel framework for humanoid whole-body control that achieves superior tracking accuracy, stability, and adaptability through automated dataset filtering, a generalist-specialist training pipeline, and a decoupled keypoint-velocity tracking strategy. Experiments show that Exbody2 outperforms prior methods by balancing feasibility and diversity in dataset selection, enabling more robust whole-body tracking, and improving generalization across diverse motion tasks. The specialist finetuning further refines performance for high-precision tasks, demonstrating the effectiveness of a structured pretrain-finetune paradigm. These contributions push the boundaries of humanoid motion control, paving the way for more expressive and stable real-world deployments.

\section{Limitations}
Although our \emph{generalist} policy already provides broad coverage of diverse motions, it does not capture the fine-grained precision of each specialist policy within a single framework. One limitation of our approach is the inability to seamlessly recombine policies that have been fine-tuned for specific motion groups. This constraint reduces the flexibility of switching between different policies within a single tracking session, as each specialist policy remains focused on a particular subset of motions. Consequently, transitions between different motion types may not be handled as smoothly or efficiently as desired. 

\emph{In principle, unifying the broad coverage of the generalist policy with the high-accuracy tracking of individual specialists could provide the best of both worlds.} Addressing this limitation by incorporating a dynamic policy-integration mechanism—where specialized policies can be adaptively blended or switched in real time—could significantly improve overall tracking precision, adaptability, and robustness, especially when dealing with complex, multi-modal motion sequences.

\clearpage
{
    \bibliographystyle{plainnat}
    \bibliography{references}
}
\clearpage

\newpage

\appendix
\section{Environments}

\begin{figure*}[htbp]
    \centering
    \includegraphics[width=\textwidth]{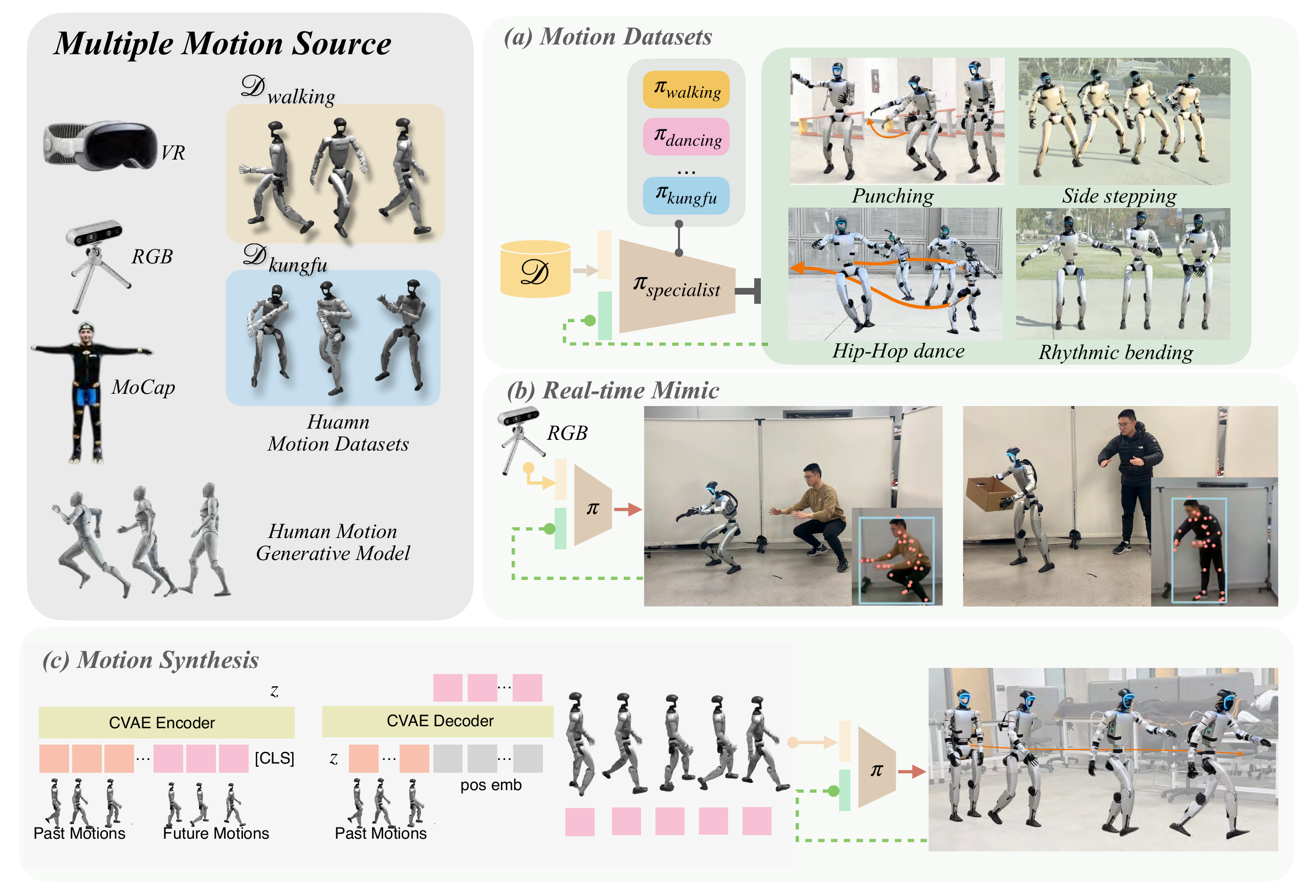}
    \caption{Illustration of ExBody2’s multi-source application, demonstrating how VR, RGB, motion capture, and generative models can be combined to produce diverse humanoid behaviors. 
    \textbf{(a) Motion Datasets:} specialized policies (e.g., kung fu, dancing) finetuned on specialist motion datasets. 
    \textbf{(b) Real-time Whole-body Mimic:} real-time replication of human motions from monocular RGB via HybrIK. 
    \textbf{(c) Motion Synthesis:} a CVAE-based approach for extended and varied motion generation. 
Experiments demonstrate ExBody2’s capability to seamlessly integrate multiple motion sources in both simulation and real-world scenarios.}
    \label{fig:multi-source}
\end{figure*}

\subsection{Real-world Deployment}
Our real robot employs a Unitree G1 platform, with an onboard Jetson Orin NX acting as the primary computing and communication device. The control policy receives motion-tracking target information as input, computes the desired joint positions for each motor, and sends commands to the robot's low-level interface. The policy’s inference frequency is set at 50 Hz. The commands are sent with a delay kept between 18 and 30 milliseconds. The low-level interface operates at a frequency of 500 Hz, ensuring smooth real-time control. The communication between the control policy and the low-level interface is realized through LCM (Lightweight Communications and Marshalling)~\cite{Huang2010LCMLC}.

\subsection{State Space Definition}
In this section, we provide detailed information on the state space used for policy training, including proprioceptive states, privileged information, and motion tracking targets.

\noindent\textbf{Robot Proprioceptive States.} The robot proprioceptive states for the teacher and the student policy can be found in Table~\ref{tab:proprio_state}. Note that the student policy is trained on longer history length compared to the teacher, as it cannot observe privileged information but have to learn from a longer sequence of past observations. \\
\textbf{Privileged Information.} The teacher policy leverages privileged information to obtain accurate motion-tracking performance. The complete information about the privileged states is listed in Table~\ref{tab:priv_info}. \\
\textbf{Tracking Target Information.} Both the teacher policy and student policy also take the motion tracking goal as part of their observations, which consists of the keypoint positions, DoF (joint) positions, as well as root movement information. The detailed components of the motion tracking target can be found in Table~\ref{tab:tracking_info}.\\
\textbf{Action Space.} The action is the target position of joint proportional derivative (PD) controllers, which is 23 dimensions for Unitree G1.

\begin{table}[htbp!]
    \centering
    \begin{tabular}{@{}ll@{}}
        \toprule
        \textbf{State} & \textbf{Dimensions} \\ \midrule
        DoF position & 23 \\
        DoF velocity & 23 \\
        Last Action & 23 \\
        Root Angular Velocity & 3 \\
        Roll & 1 \\
        Pitch & 1 \\
        Yaw & 1 \\
        \midrule
        \textbf{Total Dim} & \textbf{75*10} \\
        \bottomrule
    \end{tabular}
    \caption{Proprioceptive states used in Exbody2. The rotation information is from IMU. 10 is the length of the history proprioception}
    \label{tab:proprio_state}
\end{table}
\begin{table}[htbp!]
\centering
\begin{tabular}{ll}
    \toprule
    \textbf{State} & \textbf{Dimensions} \\ \midrule
    DoF Difference & 23 \\
    Keybody Difference & 36 \\
    Root velocity & 3 \\
    \midrule
    \textbf{Total dim } & \textbf{62} \\ 
    \bottomrule
\end{tabular}
\caption{Privileged information used in Exbody2.}
\label{tab:priv_info}

\end{table}
\begin{table}[htbp!]
    \centering
    \begin{tabular}{@{}ll@{}}
        \toprule
        \textbf{State} & \textbf{Dimensions} \\ \midrule
        DoF position & 23 \\
        Keypoint position & 36 \\
        Root Velocity & 3 \\
        Root Angular Velocity & 3 \\
        Roll & 1 \\
        Pitch & 1 \\
        Yaw & 1 \\
        Height & 1 \\ 
        \midrule
        \textbf{Total dim} & \textbf{69} \\ 
        \bottomrule
    \end{tabular}
    \caption{Reference information used in Exbody2.}
    \label{tab:tracking_info}
\end{table}

\section{Model and Training Details}
\subsection{Baseline Implementation}
\noindent\textbf{Exbody}~\cite{cheng2024express}: The implementation of Exbody is consistent with the original Exbody design, tracking only the upper-body keypoints and joint positions. For the Unitree G1 robot, we extended Exbody to tracking the all three dofs of the waist, while keeping other aspects identical. The key differences between Exbody and our method are as follows: Exbody focuses solely on upper-body motion tracking, does not utilize a teacher-student structure, uses a history length of only 5, and performs tracking entirely with local keypoints.\\
\textbf{Exbody$^\dagger$}: Exbody\textsuperscript{\(\dagger\)} is the full-body version of Exbody. It maintains most aspects of the original Exbody design but tracks the entire body's keypoints and joint positions instead of just the upper body.\\ 
\textbf{OmniH2O$^*$}~\cite{he2024omnih2o}: The main difference between OmniH2O\textsuperscript{*} and our method lies in the training phase. Specifically, OmniH2O\textsuperscript{*} does not use the robot's velocity as privileged information and relies solely on global tracking during training. For fairness, while OmniH2O\textsuperscript{*} retains its original training method, we adapted it during testing to use local keypoints for evaluating tracking accuracy. Apart from this, we ensured that the observation space and reward design were consistent with the original OmniH2O implementation.

\subsection{Policy Training Hyper-parameters}
Exbody2 adopts a teacher-student training framework. The teacher policy is trained with standard PPO~\cite{schulman2017proximal} algorithm on privileged information, tracking target and proprioceptive states. The student policy is trained with Dagger~\cite{dagger} without privileged information, but using longer history. For both teacher and student policies, we concatenate the corresponding inputs and feed them into MLP layers for policy learning. We provide the detailed training hyper-parameters for our teacher and student policy in Table~\ref{tab:teacher-student}. 
\begin{table}[htbp!]
\centering
\begin{tabular}{cc}
\toprule
Hyperparameter
     &  Value \\
     \midrule
        Optimizer & Adam \\
        $\beta_1, \beta_2$ & 0.9, 0.999 \\ 
        Learning Rate   & $1e^{-4}$\\
        Batch Size & 4096 \\
        \midrule{\textbf{Teacher Policy}} \\
        Discount factor ($\gamma$) & 0.99 \\
        Clip Param & 0.2 \\
        Entropy Coef & 0.005 \\
        Max Gradient Norm & 1 \\
        Learning Epoches & 5 \\
        Mini Batches & 4 \\
        Value Loss Coef & 1 \\
        Entropy Coef & 0.005 \\
        Value MLP Size & [512, 256, 128] \\
        Actor MLP Size & [512, 256, 128] \\
        \midrule{\textbf{Student Policy}} \\
        Student Policy MLP Size & [1024, 1024, 512] \\
        \bottomrule
\end{tabular}
\caption{Hyperparameters related to the teacher and student policy's training.}
\label{tab:teacher-student}
\end{table}

\subsection{Reward Design}
In the main paper, we partially introduced our tracking-based reward design. Additionally, our reward also contains other penalties and regularization terms. The regularization reward components and their weights to compute the final rewards are introduced in Table~\ref{tab:reward_components}. The final reward combines both the regularization with the tracking-based reward to train a robust RL policy. \\
\begin{table}[thbp!]
    \centering
    \begin{tabular}{lll}
        \toprule
        \textbf{Term} & \textbf{Expression} & \textbf{Weight} \\ 
        \midrule
        DoF position limits & $\mathbb{1}(d_t \notin [q_{\min}, q_{\max}])$ & $-10$ \\
        DoF acceleration & $\| \ddot{d}_t \|_2^2$ & $-3e^{-7}$ \\
        DoF error & $\| d_t - d_0 \|_2^2$ & $-0.1$ \\
        Action rate & $\| a_t - a_{t-1} \|_2^2$ & $-0.1$ \\
        Feet air time & $T_{\text{air}} - 0.5$ & $10$ \\
        Feet contact force & $\| F_{\text{feet}} \|_2^2$ & $-0.003$ \\
        Stumble & $\mathbb{1}(F_{\text{feet}}^x > 5 \times F_{\text{feet}}^z)$ & $-2$ \\
        Waist roll pitch error & $\| p^{\text{wrp}}_t - p^{\text{wrp}}_0 \|_2^2$ & $-0.5$ \\
        Ankle Action & $\| a^{\text{ankle}}_t \|_2^2$ & $-0.1$ \\
        \midrule
    \end{tabular}
    \caption{Regularization rewards for preventing undesired behaviors for sim-to-real transfer and refined motion.}
    \label{tab:reward_components}
\end{table}

\begin{table*}[htbp]
\centering
\small
\begin{tabularx}{\textwidth}{l X X X X X X X X}

\toprule
\multicolumn{2}{c}{} & \multicolumn{7}{c}{\textbf{Metrics}} \\
\cmidrule(r){3-9}
\textbf{\small Training Dataset} & \textbf{\small In dist.} & $E_{\text{vel}} \downarrow$ & $E_{\text{mpkpe}} \downarrow$ & $E_{\text{mpkpe}}^{\text{upper}} \downarrow$ & $E_{\text{mpkpe}}^{\text{lower}} \downarrow$ & $E_{\text{mpjpe}} \downarrow$ & $E_{\text{mpjpe}}^{\text{upper}} \downarrow$ & $E_{\text{mpjpe}}^{\text{lower}} \downarrow$ \\
\midrule

\rowcolor{lightgray}
\multicolumn{9}{l}{\footnotesize \textbf{(a) Eval. on $\mathcal{D}_{50}$}} \\
\midrule
$\mathcal{D}_{50}$ & \checkmark & \textbf{0.1375} & \textbf{0.0627} & \textbf{0.0571} & \textbf{0.0682} & \textbf{0.0753} & \textbf{0.0626} & \textbf{0.0928} \\
\midrule
$\mathcal{D}_{250}$ & \checkmark & 0.1454 & 0.0669 & 0.0600 & 0.0738 & 0.0870 & 0.0689 & 0.1119 \\

\midrule
$\mathcal{D}_{\text{CMU}}$ & \checkmark & 0.1543 & 0.0767 & 0.0649 & 0.0885 & 0.1099 & 0.0854 & 0.1437 \\
\midrule[0.5pt]

\rowcolor{lightgray}
\multicolumn{9}{l}{\footnotesize \textbf{(b) Eval. on $\mathcal{D}_{\text{CMU}}$}} \\
\midrule
$\mathcal{D}_{50}$ & \ding{55} & 0.3509 & 0.1076 & 0.1074 & 0.1076 & 0.1338 & 0.1285 & 0.1410 \\
\midrule
$\mathcal{D}_{250}$ & \ding{55} & 0.2834 & \textbf{0.1048} & \textbf{0.1021} & \textbf{0.1073} & \textbf{0.1148} & \textbf{0.1012} & \textbf{0.1335} \\
\midrule
$\mathcal{D}_{\text{CMU}}$ & \checkmark & \textbf{0.2622} & 0.1071 & 0.1036 & 0.1110 & 0.1291 & 0.1129 & 0.1512 \\
\midrule[0.5pt]

\rowcolor{lightgray}
\multicolumn{9}{l}{\footnotesize \textbf{(c) Eval. on $\mathcal{D}_{\text{ACCAD}}$}} \\
\midrule
$\mathcal{D}_{50}$ & \ding{55} & 0.4226 & 0.1277 & 0.1210 & 0.1330 & 0.1720 & 0.1618 & 0.1861 \\
\midrule
$\mathcal{D}_{250}$ & \ding{55} & 0.3533 & \textbf{0.1234} & \textbf{0.1141} & \textbf{0.1315} & \textbf{0.1421} & \textbf{0.1223} & \textbf{0.1692} \\
\midrule
$\mathcal{D}_{\text{CMU}}$ & \ding{55} & \textbf{0.3452} & 0.1267 & 0.1146 & 0.1381 & 0.1780 & 0.1635 & 0.1979 \\
\bottomrule
\end{tabularx}
\caption{Dataset Ablation Study: Evaluation on $\mathcal{D}_{50}$, $\mathcal{D}_{\text{CMU}}$, and $\mathcal{D}_{\text{ACCAD}}$ datasets with models trained on various datasets. Statistically significant results are highlighted in bold across 5 random seeds.}

\label{tab:dataset_ablation}
\end{table*}

\section{Empirical Analysis of Dataset Selection}
\label{sec:dataset-analysis}

In the main paper, we propose a \textbf{Feasibility–Diversity Principle}, which posits that a good motion dataset for humanoid tracking must be:
\begin{enumerate}
    \item \emph{Diverse enough} (especially in upper-body movements) to ensure the learned policy can generalize beyond very simple or repetitive actions.
    \item \emph{Feasible enough} that lower-body motions do not exceed the robot’s mechanical limits, avoiding extreme samples (e.g., tumbling, handstands) that hamper training.
\end{enumerate}
To illustrate how we arrived at this principle, We manually design three datasets of varying sizes, where the largest being the complete CMU dataset. The remaining datasets, with sizes 50, and 250, are subsets of the CMU dataset, each constructed with different levels of action diversity:
\begin{itemize}
    \item \textbf{50-action dataset} (\(\mathcal{D}_{50}\)): 
    A minimal set containing only fundamental and mostly static actions (e.g., standing, simple walking). 
    While highly feasible, it lacks diversity in both upper and lower limb motions.
    
    \item \textbf{250-action dataset} (\(\mathcal{D}_{250}\)): 
    A moderate-sized set extending \(\mathcal{D}_{50}\) with additional upper-limb variations (e.g., arm gestures, some dance moves) and moderately dynamic lower-body actions (e.g., running, mild jumps). 
    Crucially, it avoids highly extreme motions that are difficult for the robot to replicate.

    \item \textbf{Full CMU dataset} (\(\mathcal{D}_{CMU}\)): 
    The complete CMU motion-capture repository of 1{,}919 sequences, including extreme movements like push-ups, rolling on the ground, and somersaults. 
    Although highly diverse, it contains many infeasible actions that can introduce significant training noise.
\end{itemize}

We train separate policies with our Exbody2 framework on each dataset above and test them on three different evaluation sets:
\begin{enumerate}
    \item \(\mathcal{D}_{50}\) (in-distribution for the simplest data).
    \item \(\mathcal{D}_{CMU}\) (the full, more complex dataset).
    \item \(\mathcal{D}_{ACCAD}\), an out-of-distribution set containing actions not found in any of the training subsets.
\end{enumerate}

Table~\ref{tab:dataset_ablation} summarizes our findings:
\begin{itemize}
    \item \textbf{Evaluation on \(\mathcal{D}_{50}\)}: 
    Policies trained on \(\mathcal{D}_{50}\) unsurprisingly achieve the highest tracking accuracy for \textit{in-distribution} actions, as reflected in metrics across all categories. This suggests that additional data does not necessarily benefit in-distribution tasks. While the policy trained on \(\mathcal{D}_{250}\) performs similarly to \(\mathcal{D}_{50}\), the policy trained on \(\mathcal{D}_{CMU}\) exhibits a substantial drop in tracking accuracy.
    
    \item \textbf{Evaluation on \(\mathcal{D}_{CMU}\)}: Policies trained on \(\mathcal{D}_{250}\) achieve the best performance on \(\mathcal{D}_{CMU}\), surpassing those trained on the full \(\mathcal{D}_{CMU}\) dataset. Due to the limited diversity of the \(\mathcal{D}_{50}\) dataset, especially in upper limb movements, the \(\mathcal{D}_{50}\)-trained policy struggles to maintain high accuracy for out-of-distribution actions. Unexpectedly, the \(\mathcal{D}_{250}\)-trained policy generalizes better than the one trained on \(\mathcal{D}_{CMU}\). This result underscores that noisy datasets degrade policy performance, as the policy may expend unnecessary effort on tracking infeasible actions, lowering the accuracy of feasible actions.

    \item \textbf{Evaluation on \(\mathcal{D}_{ACCAD}\)}: This experiment further emphasizes the importance of clean datasets. Here, the ACCAD dataset (\(\mathcal{D}_{ACCAD}\)) consists of actions that are entirely not in the training data. The policy trained on \(\mathcal{D}_{250}\) outperforms the others, achieving the best tracking accuracy. Additionally, the \(\mathcal{D}_{250}\) and \(\mathcal{D}_{CMU}\)-trained policies perform relatively well in velocity tracking. However, the \(\mathcal{D}_{50}\)-trained policy suffers from substantial tracking errors, suggesting the limitations of a small, simple dataset in handling unseen data.
\end{itemize} 

\begin{figure*}[htbp]
    \centering
    \includegraphics[width=\textwidth]{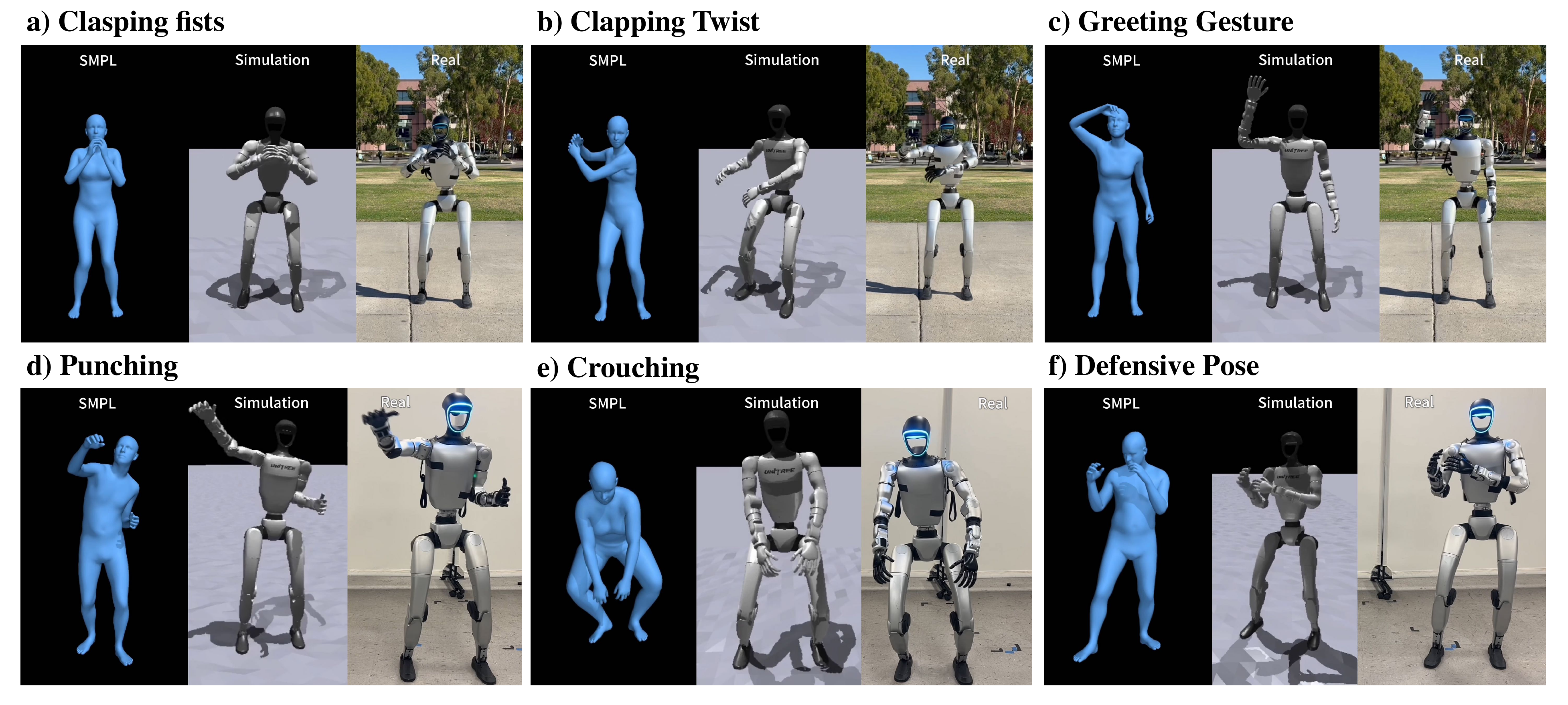}
    \caption{Sim-to-real experiment results showcasing diverse motions across SMPL, simulation, and real-world environments. Examples include: (a) Clasping Fists, (b) Clapping Twist, (c) Greeting Gesture, (d) Punching, (e) Crouching, and (f) Defensive Pose.}
    \label{fig:result_figure}
\end{figure*}
In conclusion, these results validate the core insight behind our \textbf{Feasibility–Diversity Principle}. A small dataset (\(\mathcal{D}_{50}\)) is indeed easy for the policy to master but lacks sufficient variety to generalize well. On the other hand, a fully unfiltered large dataset (\(\mathcal{D}_{CMU}\))—while highly diverse—contains many motions well beyond the robot’s capabilities, introducing detrimental noise. The \(\mathcal{D}_{250}\) subset thus provides the best balance between feasible lower-body motions and diverse upper-body actions, enabling our policy to learn robust and expressive whole-body control.

\section{Policy Ablation and Additional Results}
\label{sec:policy-ablation}

\begin{table*}[t]
\centering
\small

\begin{tabularx}{\textwidth}{l X X X X X X X X}

\toprule
\textbf{Method} & $E_{\text{vel}} \downarrow$ & $E_{\text{mpkpe}} \downarrow$ & $E_{\text{mpkpe}}^{\text{upper}} \downarrow$ & $E_{\text{mpkpe}}^{\text{lower}} \downarrow$ & $E_{\text{mpjpe}} \downarrow$ & $E_{\text{mpjpe}}^{\text{upper}} \downarrow$ & $E_{\text{mpjpe}}^{\text{lower}} \downarrow$ \\
\midrule

\rowcolor{lightgray}
\multicolumn{8}{l}{\textbf{(a) History Length Ablation}} \\
\midrule
Exbody2-History10 (Ours) & 0.2930 & \textbf{0.1000} & 0.0960 & \textbf{0.1040} & \textbf{0.1079} & \textbf{0.0953} & \textbf{0.1253} \\
Exbody2-History0  & 0.4151 & 0.1047 & 0.1010 & 0.1081 & 0.1119 & 0.0986 & 0.1303 \\
Exbody2-History25 & 0.2950 & 0.1032 & 0.0984 & 0.1078 & 0.1128 & 0.0965 & 0.1351 \\
Exbody2-History50 & \textbf{0.2648} & 0.1004 & \textbf{0.0956} & {0.1051} & 0.1114 & 0.0967 & 0.1317 \\
Exbody2-History100& 0.3242 & 0.1063 & 0.1001 & 0.1122 & 0.1225 & 0.1050 & 0.1466 \\
\midrule

\rowcolor{lightgray}
\multicolumn{8}{l}{\textbf{(b) DAgger Ablation}} \\
\midrule
Exbody2(Ours) &\textbf{0.2930} & \textbf{0.1000} &\textbf{0.0960} & \textbf{0.1040 }& \textbf{0.1079} & \textbf{0.0953} & \textbf{0.1253} \\
Exbody2-w/o-DAgger & 0.4195 & 0.1150 & 0.1106 & 0.1198 & 0.1496 & 0.1416 & 0.1607 \\
\bottomrule
\end{tabularx}
\caption{Self Ablation Study: Evaluation of different configurations of our method on dataset $\mathcal{D}_{CMU}$. The table is divided into two parts: (a) History Length Ablation and (b) DAgger Ablation.}
\label{tab:self_ablation}
\end{table*}

\subsection{Ablation on Policy Training}
We conduct ablation studies on our policy design to highlight the effectiveness of both (i) the history length for the student policy and (ii) the teacher–student (DAgger) distillation.

\paragraph{History length.}
We test student policies trained with different history lengths in Table \ref{tab:self_ablation} (a). When no extra history is used, the policy struggles to learn effectively. Among the non-zero history lengths, most policies perform similarly while the history length of 10 yields the best results, which is used by us in the main experiments. Longer history lengths increase the difficulty of fitting the privileged information, ultimately reducing tracking performance.

\paragraph{Teacher–student distillation.}
Table~\ref{tab:self_ablation} (b) shows that removing DAgger-style distillation severely degrades performance. 
Without privileged velocity guidance, the student policy must learn velocity tracking directly from raw observations, making it harder to track fast or dynamic motions accurately.

\subsection{Distribution-Guided Threshold Selection}
\begin{figure}[p]
    \centering
    \includegraphics[width=\linewidth]{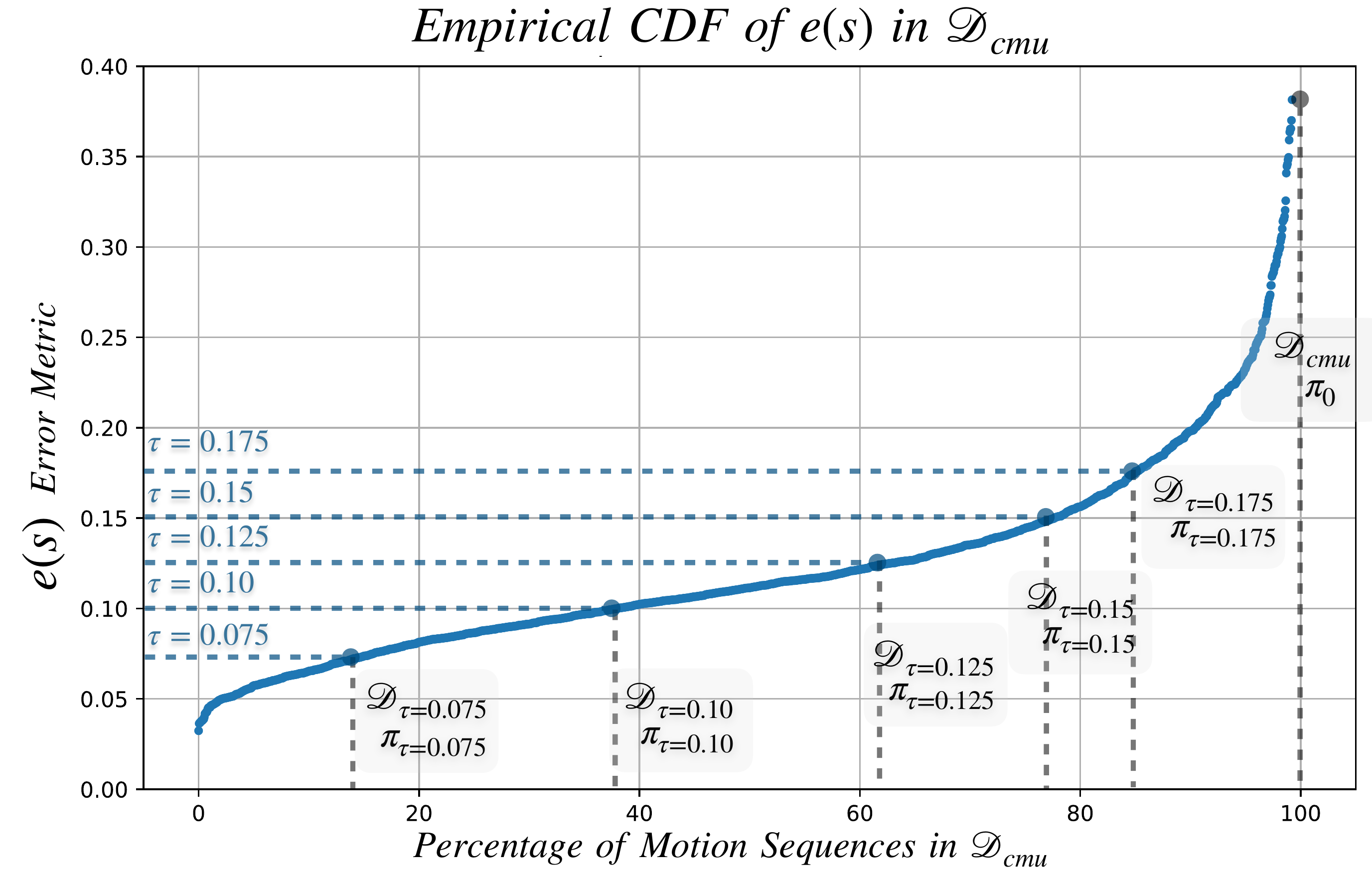}
    \caption{Empirical CDF of the base policy’s error metric $e(s)$ on the entire $\mathcal{D}_{\mathrm{CMU}}$ dataset. 
    The horizontal axis indicates the percentile of motion sequences from $0\%$ (lowest error) to $100\%$ (highest error), 
    while the vertical axis shows $e(s)$. 
    We overlay dashed horizontal lines at key thresholds ($\tau = 0.075, 0.10, 0.125, 0.15, 0.175$) to illustrate how 
    we systematically determine feasible versus unfeasible motions based on the empirical distribution.}
    \label{fig:cdf}
\end{figure}

To choose filtering thresholds in a principled manner, we first analyze the error distribution of the base policy across the entire dataset. Figure~\ref{fig:cdf} presents the empirical cumulative distribution of $e(s)$, with the x-axis indicating the percentile of motion sequences (from lowest to highest error) and the y-axis displaying the corresponding error value. 

We derive thresholds directly from the empirical distribution, ensuring a data-driven rather than arbitrary cutoff. 
Smaller thresholds yield mostly lower-body motions with limited dynamics, while gradually increasing the threshold 
admits more dynamic behaviors. Higher thresholds include samples with excessive errors that could degrade policy learning. Consequently, we select 
\(\tau = 0.075,\,0.10,\,0.125,\,0.15,\,0.175\) to filter the dataset into subsets of varying feasibility and diversity. 
This data-driven approach aligns with our \textbf{feasibility-diversity principle}, yielding balanced subsets that 
support robust policy learning.

\subsection{Real-world Results Visualization}
Figure~\ref{fig:result_figure} illustrates how ExBody2 successfully replicates various motions in both simulation and real-world settings. 
We align each frame’s pose from (i) the reference SMPL animation, (ii) our simulated humanoid robot, and (iii) the real robot deployment. 
These snapshots confirm that our learned policy retains high fidelity to the target motion, including lower-body poses critical for balance. Additional results can be viewed in the supplementary video.

\section{ExBody2’s Multi-source Demonstration}
\label{sec:multisource}
One key advantage of ExBody2 is its flexibility in handling multiple motion sources. 
In the main text, we focus primarily on motion capture data (i.e., offline datasets).
Below, we highlight two other sources—\emph{Real-time Whole-body Mimic} (RGB-based) and \emph{Motion Synthesis} (latent generative model)—that can drive ExBody2 for more interactive and long-horizon tasks. 
Figure~\ref{fig:multi-source} visually summarizes these capabilities alongside possible VR or IMU-based streams.

\subsection{Real-time Whole-body Mimic (RGB Input)}
\label{sec:realtime-mimic}
We implement a real-time tracking pipeline that uses only \textit{monocular RGB input} to mimic human movements. 
Our system first applies the HybrIK algorithm~\cite{li2021hybrik} to extract 3D human poses from each image frame. 
We then retarget this sequence of poses to the robot’s kinematic structure and feed it into the ExBody2 whole-body policy. 
Because our policy is trained to be robust to partial or noisy signals, it can accommodate real-time streaming of 3D keypoints and still maintain stable lower-body tracking. 
Figure~\ref{fig:multi-source} (b) demonstrates a user controlling the robot to lift and carry an object, showcasing responsive teleoperation.

Relying on monocular pose estimates is more lightweight than requiring a full-body Mocap or multi-camera setup. 
Although the 3D pose can be less accurate than multi-view solutions, our control policy’s robust design helps it remain stable even under potential keypoint noise.

\subsection{Motion Synthesis for Extended Behaviors}
\label{sec:motion-synthesis}
We further incorporate a \textit{Conditional Variational Autoencoder (CVAE)} to generate new motion segments based on a short sequence of past motions, as Figure~\ref{fig:multi-source} (c) illustrated.
During inference, each latent code $z$ is sampled (or set to the prior mean) to produce new motion trajectories that seamlessly continue from the current pose. 
Unlike naive random sampling, the CVAE ensures continuity by conditioning on past pose context and penalizing abrupt transitions with a smoothness loss.

\paragraph{Training details.}
The CVAE is trained on a broad set of humanoid motion clips, optimizing a reconstruction loss plus KL-divergence for the latent space. 
We also add a small penalty for high-frequency velocity changes, improving the realism of the generated motions.

\paragraph{Integration with ExBody2.}
The generated motion frames are retargeted in exactly the same way as a regular Mocap clip, so the policy sees no difference. 
This allows the robot to perform extended, varied sequences—e.g., spontaneously transitioning from walking to an upper-body gesture—without needing to rely on a fixed database of motion capture clips.

\end{document}